\documentclass[letterpaper]{article}
\usepackage{aaai}
\usepackage{times}
\usepackage{helvet}
\usepackage{courier}
\usepackage{xcolor}
\usepackage[hidelinks]{hyperref}

\begin{document}
%
\title{Examining the Role of Clickbait Headlines to Engage Readers with Reliable Health-related Information}

\author{Sima Bhowmik \\ School of Journalism and New Media \\ University of Mississippi \\ srbhowmi@go.olemiss.edu \And
Md Main Uddin Rony \\ College of Information Studies \\ University of Maryland, College Park \\ mrony@umd.edu \And
Md Mahfuzul Haque \\ Philip Merrill College of Journalism \\ University of Maryland, College Park \\ mhaque16@umd.edu
\AND
Kristen Alley Swain \\ School of Journalism and New Media\\ University of Mississippi \\ kaswain@olemiss.edu \And
Naeemul Hassan \\ Philip Merrill College of Journalism\\ University of Maryland, College Park  \\ nhassan@umd.edu}

\maketitle

\begin{abstract}
\begin{quote}
Clickbait headlines are frequently used to attract readers to read articles. Although this headline type has turned out to be a technique to engage readers with misleading items, it is still unknown whether the technique can be used to attract readers to reliable pieces. This study takes the opportunity to test its efficacy to engage readers with reliable health articles. A set of online surveys would be conducted to test readers' engagement with and perception about clickbait headlines with reliable articles. After that, we would design an automation system to generate clickabit headlines to maximize user engagement.   
\end{quote}
\end{abstract}
\section{Introduction}
\label{sec-intro}

Ordinary people are increasingly going on the Internet to seek health-related information. A Pew Research Center study found that 72\% of adult internet users in the United States searched online for information about a range of health issues \cite{pew2014}. Both true and false information is vying for readers' attention on the online platforms. Research has found that misleading health-related information grabs more attention than evidence-based reports \cite{forster2017,raphael2019healthnews}.

When ordinary people come across misleading health-related information, many of them are likely to believe in it and take action accordingly \cite{shapiro2018}, a problem health professionals have constantly been facing. The trend offers room for more research to find out strategies to engage readers with reliable health-related information. 

To this end this paper plans to apply a novel approach of examining the use of clickbait headlines to engage readers with reliable health news. Clickbait, a writing and linguistic technique applied to write a headline which aims to trick readers into clicking links \cite{chakraborty2016stop}, has widely been used in unreliable health-related articles. For example, \cite{raphael2019healthnews} found a vital role clickbait headlines played in attracting readers to click and share content on social media. The author analyzed the credibility of the 100 most popular health articles shared on social media in 2018 to find that three quarters of the top 10 shared articles were either misleading or included some false information. Clickbait headlines were the prominent feature of these viral articles. \cite{dhoju2019differences} also found similar results that unreliable outlets used more clickbait headlines to spread health news online than reliable outlets.

Although clickbait headlines have predominantly been used in unreliable articles, legacy news media have also been using the technique to engage readers. Clickbait itself as a practice may not necessarily be objectionable; however, when it is used to lure readers to something that fails to keep its promise, then its usage becomes questionable.

As the objective of this paper is to increase reader engagement with reliable health-related information, it will examine how audiences would respond to clickbaits when it is used with reliable articles. To test engagement with health messages and promotion, previous studies have examined the use of social networking sites like Facebook and Twitter \cite{neiger2013evaluating,thrul2015smoking,neiger2013use,bhattacharya2017social}, but no studies were found that applied the clickbait technique to examine whether it helps increase engagement with reliable health-related information.

Engagement in this study has been seen as a two-way process between an outlet and its audiences where readers would share, spend time, and comment \cite{maksimainen2017improving}.

To examine the applicability of clickbaits to increase engagement, this study would employ three phases in its research design in which engagement with clickbaits would be tested through an online experiment survey in the first phase. Participants in the second phase would be asked about their perception of clickbaits. In the last phase automation would be applied to generate clickbait headlines.

The results of this study would contribute to increase engagement with reliable health-related information. Its results could be helpful for health agencies like the Centers for Disease Control and Prevention (CDC) which aims to engage more people with reliable health-related information online to help them take informed decisions. 

Its results also would potentially be applied to other domains like environment, politics, and so on.

\section{Research Plan}
\label{sec-research-plan}
In order to understand how clickbait headlines work to engage readers with reliable health-related information, we would employ the following three steps.

\subsection{First Step - Engagement with clickbaits}
In this step we would conduct an online experiment survey in which we would provide the participants with some health-related articles with headlines both in traditional format and clickbait to measure engagement. \cite{chakraborty2016stop,chen2015misleading} studied different stylometric techniques used in clickbait articles which we would incorporate in designing different clickbait headlines for a single article. Moreover, we would also explore different clickbait types identified by \cite{biyani20168} to find out the most effective type of technique for better engagement. Before providing participants with articles, we would conduct a pre-experiment survey to find information about demographics, age, education, and political views. In the post-experiment survey, we would ask participants after reading the articles whether they like to click, share, give reaction, discuss with others both online and off-line. 

Participants in the experiment survey would be recruited from the United States using the online crowdsourcing website, Amazon Mechanical Turk (MTurk). The topics of the health-related articles for the experiment would be chosen carefully to avoid confirmation bias; for example, topics related to vaccination would not be given, because many participants are likely to have some prior confirmation bias. Instead We would include topics such as nutrition, obesity, smoking, tobacco use, and so on.   

\subsection{Second Step: Perception about clickbaits}

In this step we would measure how participants perceive the news articles with clickbait headlines. We would ask readers whether they believe in the articles or not, the types of emotion it creates such as trust, fear, distrust, curiosity, excitement, disgust. We would also ask the participants about the usefulness of the articles.

     
\subsection{Third Step: Automated clickbait headline generation}
Our third step will focus on automatically generating a clickbait headline based on the content of the article which can ensure better user engagement. We plan to explore deep learning based natural text generation (NTG) models for headline generation. Based on the survey results, we will identify the suitable types and characteristics of the clickbait headlines that can ensure more engagement. The effective types of clickbait may vary with the topics of the health news, users' demographic properties, and also with the temporal and local dependencies. So, the factors which are influential in determining ideal clickbait headline can be considered as features for the generative models. 
There are several challenges involved in this task.
\subsubsection{Clickbait type and characteristic selection}
Biyani et al. \cite{biyani20168} identified 8 types of clickbait and all of which might not be appropriate for the health news. For example, the ambiguous or factually wrong headlines might mislead the readers which can impact negatively. Moreover, all clickbait characteristics don't have the same contribution to make a headline effective \cite{kuiken2017effective}. 
So, finding out the suitable type and characteristics of clickbaits will be challenging because the curiosity aroused by the headline may vary from topic to topic and also may depend on the users' demographic properties. Moreover, the temporal and locality characteristics may affect the efficacy of clickbait headlines to engage more users. Along with the textual content the ideal solution should also incorporate these local, temporal and user demographic properties as a feature to generate an effective headline which can drive more attention of the users. We are hopeful that the user study data collected in the first two phases would be useful in deciding these factors.

\subsubsection{Generating non-misleading clickbait headlines}
The contextual gap between the generated headline and the main content should not be wide so that users get immediate and relevant satisfaction after visiting the landing page. Otherwise, the news site may suffer a high bounce rate leading to the organization's reputation risks. Even though the generated headlines will be attractive, it's unlikely that all the readers who come across the article while surfing social media will click it to access to the full content. As most of the people perceive news just skimming through the headlines only and even they share the articles before reading the full content \cite{gabielkov2016social}, we need to make sure that the users should not get any misconception about the content from the generated headline. So, it is important that the generated headline should reflect the original content properly. Although the task is challenging, Shu et al. \cite{shu2018deep} provided an outline of generating stylized and synthetic headlines preserving certain information of the documents using deep generative model.

\subsubsection{Model Selection}
As we are planning to generate the headline from the content, it can be viewed as a sequence generation problem \cite{sutskever2014sequence}. Although deep learning-based auto encoder-decoder models can generate a sequence of texts from a source text, the models need to be tuned significantly to achieve a good performance. To overcome this problem, researchers proposed some other text generation methods based on reinforcement learning and adversarial training but all the models seem to have some limitations (e.g. suffering from exposure bias problems, gradient vanishing, mode collapse problems, etc.) \cite{lu2018neural}. To find a good performing model which will fit our problem easily can be a challenging task. But to begin with, we are particularly interested in trying Recurrent Neural Networks (RNNs) based  Variational Auto-Encoder (VAE) proposed by Shen et al \cite{shen2017style} where they developed a cross-aligned auto-encoder which can transfer the style of sentences preserving the content of original sentences. As our main purpose is to generate engaging headlines which contain different clickbait styles and reflect the original content, the suggested model is worthy to try out.

\subsubsection{Dataset Preparation}
To train our model we need a complete dataset containing clickbait of all types. Moreover, the clickbait headlines should be enriched with prominent clickbait features so that the generator model can learn them well. But in our best knowledge, such type of complete dataset is not available at this moment. Clickbait challenge dataset contains a sufficient number of clickbait and non-clickbait articles but the dataset contains all types of news, not particularly health-related articles. Moreover, the types of clickbait are also absent in the dataset. Dhoju et al. \cite{dhoju2019differences} curated a dataset of health news articles where the clickbait nature of the headline is also present. But this dataset also lacks the information of clickbait type. Yet this dataset can be a good starting point where we can manually label the clickbait type of a small dataset and explore the style transfer process outlined by Shu et al. \cite{shu2018deep}  to generate the synthesized headlines to build our desired training set. 




\section{Related Work}
\label{sec-related-work}
Previous literature has examined the practice of using clickbaits in mainstream news media. For example, \cite{palau2016reference} examined the content of four online sections of the Spanish newspaper \textit{El Pais} and identified various linguistic techniques were used in headlines of these articles such as orality markers and interaction (e.g.,
direct appeal to the reader), vocabulary and word games (e.g.,
informal language, generic or buzzwords), and morphosyntax
(e.g., simple structures).

Chartbeat, an analytics firm that provides market intelligence to media organizations, tested 10, 000 headlines from over 100 websites for their effectiveness in engaging users with content. The study examined 12 `common tropes' in headlines – a majority of them are considered clickbait techniques – and found that some of these tropes are more effective than others \cite{breaux2015you}. 

Some studies also examined the role of clickbaits in engaging readers. \cite{rony2017diving} examined the context of clickbait and non-clickbait articles and found that clickbait headlines generated more engagement than non-clickbaits. This study used headlines from different topics like politics, sports, environment, but not specifically about health. 

\cite{vincent2018studying} found that cognitive, affective, and pragmatic elements are significantly related to click clickbaits headline for healthcare personnel.

However, \cite{scacco2016investigating} found that question-based headlines lead to negative attitudes about the headline. Although this study did not consider the health-related articles, its findings would nevertheless be helpful for our study to test the efficacy of this type in the context of health-related articles. 

\cite{berger2012makes} investigated the reasons behind certain pieces such as advertisements, videos, news articles go more viral. The results indicate that positive content is more viral than negative content, but the relationship between emotion and social transmission is more complex than valence alone. Virality is partially driven by physiological arousal. Content that evokes high-arousal, positive or negative emotions, is more viral. Conversely, content that evokes low-arousal, or deactivating emotions is less viral.

\cite{clickbait2015generator} represents an automated clickbait generation tool which uses an RNN model trained on two million headlines collected from Buzzfeed, Gawker, Jezebel, Huffington Post, and Upworthy. The model is then used to produce new clickbait headlines. \cite{shu2018deep} used deep generative models to generate synthetic headlines with specific style labels and explored their utilities to help improve clickbait detection.

Some other studies also examined the automated detection of clickbait headlines using natural language processing \cite{chakraborty2016stop,anand2017we,potthast2016clickbait,thakur2016identifying}. 




\bibliographystyle{aaai}
\bibliography{main}

\end{document}